\documentclass[runningheads,a4paper]{llncs}
\usepackage[margin=1in]{geometry}
\usepackage[utf8x]{inputenc}
\usepackage[section]{placeins}
\usepackage{amssymb}
\usepackage{graphicx}
\usepackage{booktabs}
\usepackage{array}
\usepackage{url}
\usepackage{subfigure}
\urldef{\mailsc}\path|nathansh@icmc.usp.br, magali.duran@uol.com.br, sandra@icmc.usp.br|    
\newcommand{\keywords}[1]{\par\addvspace\baselineskip
\noindent\keywordname\enspace\ignorespaces#1}

\begin{document}

\mainmatter  
\title{Automatic semantic role labeling on non-revised syntactic trees of journalistic texts}

\titlerunning{Automatic semantic role labeling on non-revised syntactic trees}
\author{Nathan Siegle Hartmann \and Magali Sanches Duran \and Sandra Maria Aluísio}
\institute{Interinstitutional Center for Computational Linguistics (NILC)\\Institute of Mathematical and Computer Sciences\\University of São Paulo\\
\mailsc}
\maketitle

\begin{abstract}
Semantic Role Labeling (SRL) is a Natural Language Processing task that enables the detection of events described in sentences and the participants of these events. For Brazilian Portuguese (BP), there are two studies recently concluded that perform SRL in journalistic texts. \cite{alvamanchego2013} obtained F1-measure scores of 79.6, using the PropBank.Br corpus, which has syntactic trees manually revised; \cite{fonseca2013}, without using a treebank for training, obtained F1-measure scores of 68.0 for the same corpus. However, the use of manually revised syntactic trees for this task does not represent a real scenario of application. The goal of this paper is to evaluate the performance of SRL on revised and non-revised syntactic trees using a larger and balanced corpus of BP journalistic texts. First, we have shown that \cite{alvamanchego2013}'s system also performs better than \cite{fonseca2013}'s system on the larger corpus. Second, the SRL system trained on non-revised syntactic trees performs better over non-revised trees than a system trained on gold-standard data. 

\keywords{Semantic role labeling. Non-revised syntactic trees. Brazilian Portuguese.}
\end{abstract}

\section{Introduction}
Semantic Role Labelling (SRL) is a Natural Language Processing (NLP) task responsible for detecting events described in sentences and the participants of these events \cite{palmeretal2010}. The events are held by predicators, such as verbs and eventive names (some nouns, adjectives and adverbs) and the participants are called arguments. This work focuses on verbs.

To automatically annotate a text with semantic roles, most current SRL systems employ Machine Learning (ML) techniques. When using ML, the SRL task is generally performed on syntatic trees due to the extensive set of features that have been identified in the syntactic structure of a sentence, such as those presented by \cite{palmeretal2010}.

However, as the syntactic trees of a sentence are generated by automatic parsers and these tools are subject to errors, problematic trees are often manually revised by linguists. For Brazilian Portuguese SRL, the best performance was obtained by \cite{alvamanchego2013}, with F1-measure scores of 79.6 when annotating revised syntactic trees, without reported outcomes for non-revised trees. However, the use of corrected trees does not represent a real scenario of application. In this sense, Fonseca's work \cite{fonseca2013}, following an approach that does not use syntactic features, obtained F1-measure scores of 68.0 in Portuguese sentences. It is known that SRL systems using syntactic features perform better than those that do not use them. However, there are no SRL results for Portuguese on non-revised syntactic trees.

This work evaluates the use of revised and non-revised syntactic trees for manual and automatic SRL tasks in Brazilian Portuguese. We demonstrate that human annotation errors are directly related to annotation errors made by machines. We attribute these errors to problematic syntactic trees generated by the parser. We also show that, for a good performance in automatic SRL on non-revised syntactic trees, it is necessary to train the SRL system with the same type of data and/or invest in improving the parser used to preprocess the corpus.

In Section \ref{sec:corpus_selection}, we show the corpora collected and compiled in this work. In Section \ref{sec:manual_annotation}, we present the methodology for manual annotation of the corpus whose non-revised syntactic trees are annotated with semantic roles. In Section \ref{sec:related_works}, we present the state-of-art SRL system for Portuguese and a system whose methodology does not rely on syntactic errors. In Section \ref{sec:experiments}, we show conducted experiments and the obtained results. At last, Section \ref{sec:conclusions} presents this work’s conclusions.

\section{Selection of the Corpora}
\label{sec:corpus_selection}

To evaluate the SRL task on syntactic trees with errors, we annotated semantic roles in a new corpus compiled for this work, whose syntactic trees had not been revised, and also used a corpus annotated with semantic roles, whose syntactic trees had been manually revised, so that we could compare the results. The syntactic trees of both corpora were generated by the PALAVRAS parser \cite{bick2000}. The corpora used in this work are PropBank.Br version 1.1\footnote{Available at \url{http://nilc.icmc.usp.br/portlex/images/arquivos/propbank-br/PropBankBr_v1.1.xml.zip}.} \cite{duranaluisio2012}, referred to as PB-Br.v1 and a selection of the PLN-Br, corpus of texts from Folha de São Paulo \cite{bruckschenetal2008}, referred to as PB-Br.v2\footnote{Available at \url{http://nilc.icmc.usp.br/semanticnlp/propbankbr/pbbr-v2.html}.}.

In the PB-Br.v1 corpus, we observed that many verbs have only one annotation instance and this data sparsity is undesirable for machine learning purposes. When it comes to learning annotation of semantic roles, we have to consider three aspects: (i) which verbs are represented in the corpus; (ii) which meanings of verbs are represented in the corpus; and (iii) which syntactic alternations are represented in the corpus. Alternation is changing the order of constituents (syntactic, semantic, or both at once). For example, the passive voice is a syntactic alternation marking a semantic alternation (the patient takes the place of the agent in the syntactic subject position). Normally, the number of meanings for a single verb is associated with the amount of syntactic alternations that it admits, but there may be verbs that admit a large number of alternations for the same meaning. Our goal in compiling the PB-Br.v2 was to get as much representation as possible in the three items above with the lowest number of annotation instances, since manual annotation is costly. The sentences were divided into classes according to linguistic criteria and considering the ML, as we chose to start the annotation process with the most frequent verbs. Using corpus statistics, we found that verbs with a frequency higher than 1000 represent 90\% of verb occurrences in the corpus. We assume that they are good material for training a classifier of semantic roles, because if the classifier learns to rank well 90\% of the corpus it should have good accuracy. In addition, the highly frequent verbs, excluding auxiliary, copula verbs and verbs that require clausal complements, are probably the most polysemous in the language. More information about the selection of sentences for the PB-Br.v2 can be found in \cite{duranetal2015}\footnote{Available at \url{http://nilc.icmc.usp.br/semanticnlp/propbankbr/relat.html}.}.  

The PB-Br.v1 corpus contains 5,931 annotated instances of 3,348 sentences and the PB-Br.v2 contains 7,661 annotated instances of 7,442 sentences. We generate instances of a sentence for each verb contained in it. Furthermore, we only selected instances whose syntactic trees generated by the parser are related, i.e. all elements of the syntactic tree are connected. Section \ref{sec:manual_annotation} presents the results of the manual annotation of semantic roles on the selection made for the PB-Br.v2.

\section{Manual Annotation of the PB-Br.v2}
\label{sec:manual_annotation}

The manual annotation of semantic roles was performed on 7,661 selected instances of the PB-Br.v2, following the annotation of the PropBank project \cite{palmeretal2005}, but based on annotation guidelines\footnote{Available at \url{http://nilc.icmc.usp.br/semanticnlp/propbankbr/manual.html}.} customized to the Portuguese language enriched with wrong syntactic trees annotation process.  The Tiger XML output of the PALAVRAS syntactic parser was used, in the same format of the PB-Br.v1 corpus. Furthermore, Tiger XML syntactic trees can be processed with the SALTO tool \cite{burchardtetal2006b} that was used in annotating the corpus in question. 

A group of seven annotators and one adjudicator participated in the annotation process. The annotators were trained and received a copy of the annotation guidelines. In addition, a repository of verbs and their meanings, called Verbo-Brasil, was available during the annotation\footnote{Available at \url{http://143.107.183.175:12680/verbobrasil/}.}. The task consisted of annotating the meaning of the verb – to identify the set of expected semantic roles – and assigning semantic role labels chosen from a set of six numbered arguments (ArgN) and 12 modifiers (ArgM). The description of each role is detailed in the annotation manual. The annotation scheme followed the double-blind standard, in which two annotators annotate the same portion of instances and, afterwards, the adjudicator solves disagreements. We distribute the annotation of instances by blocks, gathering instances of the same verb in the same task. Separately, the sets of copula verbs and verbs that require clausal complements were distributed. The same annotation instance was never assigned to more than two annotators. 

After concluding the annotation, we calculated the Kappa statistics \cite{carletta1996}. As the SRL task traditionally consists of two steps: identification of arguments and classification of semantic roles of each argument, we calculated the individual Kappa for each step: 0,96 and 0,90 for copula verbs; 0,96 and 0,95 for clausal complement verbs; and 0,79 and 0,75 for verbs with no syntactic pattern. We also calculated the Kappa statistics of the identification of verb meaning: 1,0 for copula verbs, 1.0 for clausal complement verbs and 0,92 for verbs with no syntactic pattern.


The set of verbs that require clausal complements obtained important Kappa results, considered almost perfect by the Kappa scale of Landis and Koch \cite{landiskoch1977}. As the annotation of this set is very predictable and has little syntactic diversity, the identification and classification obtained almost maximum values. The annotation of copula verbs also obtained almost perfect results. In this scenario, however, there is a greater syntactic diversity, which led to a drop in the agreement of argument classification compared to the set of verbs that require clausal complements. This means that the annotators agreed in identifying that a particular syntactic node is the verb argument, but disagreed in selecting its semantic role. Finally, the set of plain verbs that present several alternations obtained a substantial Kappa that, in the used Kappa scale, is lower than that of the other annotated verb sets. For this set, we found that there was disagreement in selecting the meaning of verbs, which triggered disagreements about the detection of arguments and the selection of semantic roles. Additionally, this is the set of verbs whose annotation is far from being predictable as they present a lot of syntactic alternation, i.e., syntactic constituents may appear in different orders, which affects the distribution of roles. After analyzing human annotation, we separated the portion in which there was disagreement between the annotators and sent it to the adjudicator, who solved the disagreements and generated a final version of the annotation. Table \ref{tab:concordanciaanotacaomanualplnbr} shows the number of instances in which there was full agreement between annotators and those that had to be revised. We noted there was disagreement in over 50\% of the instances. This means that, despite the high Kappa values, most instances had to be revised. However, this does not imply a high rate of disagreement – as suggested by the Kappa values. For example, an instance is sent for adjudication even if annotators have agreed on 9 out of 10 annotated arguments and disagreed on just one.

\newcolumntype{R}[1]{>{\raggedleft\let\newline\\\arraybackslash\hspace{0pt}}m{#1}}

\begin{table}[!ht]
\center
\scriptsize
\scalebox{.91}{
\begin{tabular}{lR{2.1cm}R{2.1cm}R{2.1cm}}
\toprule
 & \textbf{Full Agreement} & \textbf{Some Agreement} & \textbf{Total}\\
\midrule
Copula verbs & 46 (64.7\%) & 25 (35.2\%) & 71\\
Verbs of clausal complements & --- & --- & 602\\
Other plain verbs & 3,130 (40,7\%)& 4,542 (59,2\%) & 7,672\\
\midrule
\textbf{Total} & \textbf{3,176} & \textbf{4,567} & \textbf{8,345}\\
\bottomrule
\end{tabular}
}
\caption{Proportion of the PB-Br.v2 annotation in which there was full agreement.}
\label{tab:concordanciaanotacaomanualplnbr}
\end{table}

It is important to note that many syntactic trees contain errors generated by the parser. As the annotation guidelines do not cover how to deal with all possible errors generated by the parser, the free interpretation of how to annotate these inconsistent trees may have contributed to the number of disagreements obtained.

\section{SRL Systems}
\label{sec:related_works}

This paper analyzes the performance of two newly developed systems to annotate semantic roles of texts written in Brazilian Portuguese. 
Fonseca \cite{fonseca2013} developed a system for annotation of semantic roles for Brazilian Portuguese, avoiding dependence on external NLP tools, such as syntactic parser. Its results on revised syntactic trees of the PB-Br.v1 corpus were a F1-measure of 68.0. 
The work of Alva-Manchego \cite{alvamanchego2013}, which uses the PB-Br.v1 corpus for training, employed a supervised approach for automatic annotation of semantic roles on syntactic trees of Brazilian Portuguese sentences. The system performance on revised syntactic trees of the PB-Br.v1 corpus was a F1-measure of 79.6.
The main difference between the systems is the use of syntactic features (the former does not use them). In addition, none of the systems evaluated their performance on non-revised syntactic trees. The contrast on the SRL performance on revised and non-revised syntactic trees has never been done for Portuguese. Yet, few studies relate the causes of errors in human annotation with the errors generated by the machine in the SRL task, and this is one of the contributions of our work.

\section{Experiments}
\label{sec:experiments}

In this section, we present the experiments developed to contrast the performance of SRL systems trained on syntactic trees revised by humans (treebanks or gold-standard data) and non-revised syntactic trees. We used Alva-Manchego’s system \cite{alvamanchego2013} in most experiments because it is the state-of-art system in SRL for Brazilian Portuguese. We also used Fonseca’s system \cite{fonseca2013} in a final experiment to contrast its performance with the one of Alva-Manchego’s system. Fonseca’s system does not use syntactic trees and, therefore, it does not suffer from the syntactic errors generated by the parser. In the scenario of the PB-Br.v2, it is interesting to note whether the errors generated by the parser are significant to cause the Alva-Manchego’s system to have a SRL performance lower than Fonseca’s.

The experiments reported below evaluate Alva-Manchego’s system on different sets of the PB-Br.v2: (i) the Agreement set, composed of instances where the annotators fully agreed, therefore, not sent for adjudication; (ii) the Adjudication set, composed of instances that were sent for adjudication; and (iii) the Full set, comprising Agreement and Adjudication.

First, we conducted a 10-fold cross-validation for the set where there was agreement between PB-Br.v2 annotators. The following results are categorized by ``corr.'' indicating correct labeling, ``excess'' indicating a mislabeled annotation, ``missed'' indicating an argument that was not selected as a candidate to annotation, ``prec.'' as precision, ``rec.'' as recall and ``F1'' as F-measure. The results in Table \ref{tab:concordanciaanotacaoplnbr} show a reasonable indication of the SRL quality, since the Agreement set has approximately 50\% of the amount of PB-Br.v1 instances used by Alva-Manchego. To contrast it with the Agreement set, we also checked the quality of the SRL on the Adjudication set. The results in Table \ref{tab:adjudicacaoanotacaoplnbr} show that the Adjudication set, in spite of being approximately 80\%  greater than the PB-Br.v1  and approximately 40\% greater than the Agreement set, does not allow an easy (or simple) automatic annotation of semantic roles. We can interpret that if humans find it hard to annotate, the machine will have the same problem – which justifies the difference of F1 scores of 6.98 on the performance of the Agreement (F1 = 72.71) and Adjudication (F1 = 65.73) sets. We can also speculate that the parsing errors, which caused different annotation interpretations to the human eye, hindered the automatic learning of the task because it increases data sparsity.

\begin{table}[!ht]
\center
\scriptsize
\scalebox{.88}{
\parbox{.50\linewidth}{
\begin{subtable}

\begin{tabular}{lrrrrrr}
\toprule
& \textbf{corr.} & \textbf{excess} & \textbf{missed} & \textbf{prec.} & \textbf{rec.} & \textbf{F1}\\
\textbf{Overall} & 3,640 & 1,401 & 1,332 & 72.23 & 73.22 & 72.71\\
\midrule
A0 & 807 & 230 & 146 & 77.98 & 84.64 & 81.09\\
A1 & 1,853 & 491 & 600 & 79.13 & 75.56 & 77.28\\
A2 & 367 & 294 & 223 & 55.48 & 62.14 & 58.42\\
A3 & 10 & 17 & 43 & 45.67 & 26.82 & 27.83\\
A4 & 2 & 1 & 15 & 20.00 & 8.33 & 11.67\\
AM-ADV & 9 & 12 & 13 & 28.00 & 26.67 & 24.56\\
AM-CAU & 2 & 4 & 16 & 13.33 & 8.33 & 9.00\\
AM-DIR & 0 & 0 & 0 & -- & -- & --\\
AM-DIS & 6 & 8 & 16 & 21.67 & 16.67 & 18.16\\
AM-EXT & 16 & 4 & 11 & 75.00 & 60.83 & 64.33\\
AM-LOC & 103 & 113 & 50 & 48.69 & 67.53 & 55.54\\
AM-MNR & 80 & 75 & 57 & 52.57 & 60.55 & 54.88\\
AM-NEG & 134 & 14 & 19 & 90.30 & 87.80 & 88.83\\
AM-PRD & 0 & 0 & 1 & 0.00 & 0.00 & 0.00\\
AM-PRP & 39 & 28 & 17 & 58.89 & 71.23 & 62.89\\
AM-REC & 0 & 0 & 0 & 0.00 & -- & --\\
AM-TMP & 212 & 110 & 105 & 65.75 & 67.07 & 66.03\\
\bottomrule
\end{tabular}

\caption{10-fold cross-validation results for the PB-Br.v2 Agreement set.}
\label{tab:concordanciaanotacaoplnbr}
\end{subtable}
}}
\quad\quad
\scalebox{.88}{
\parbox{.52\linewidth}{
\begin{subtable}

\begin{tabular}{lrrrrrr}
\toprule
& \textbf{corr.} & \textbf{excess} & \textbf{missed} & \textbf{prec.} & \textbf{rec.} & \textbf{F1}\\
\textbf{Overall} & 7,164 & 4,001 & 3,469 & 64.16 & 67.37 & 65.73\\
\midrule
A0 & 1,873 & 491 & 377 & 79.26 & 83.24 & 81.18\\
A1 & 2,703 & 1,208 & 1,142 & 69.12 & 70.29 & 69.68\\
A2 & 489 & 708 & 517 & 40.82 & 48.76 & 44.32\\
A3 & 14 & 36 & 82 & 28.39 & 14.14 & 18.00\\
A4 & 8 & 5 & 23 & 36.00 & 19.05 & 23.00\\
AM-ADV & 157 & 134 & 119 & 54.37 & 57.50 & 55.49\\
AM-CAU & 47 & 67 & 75 & 42.80 & 38.17 & 39.43\\
AM-DIR & 0 & 5 & 13 & 0.00 & 0.00 & 0.00\\
AM-DIS & 150 & 76 & 172 & 66.76 & 46.85 & 54.69\\
AM-EXT & 56 & 35 & 43 & 62.70 & 58.19 & 58.55\\
AM-LOC & 413 & 374 & 161 & 52.45 & 72.60 & 60.79\\
AM-MNR & 200 & 281 & 219 & 41.81 & 47.78 & 44.48\\
AM-NEG & 226 & 22 & 39 & 90.92 & 86.16 & 88.10\\
AM-PRD & 46 & 125 & 124 & 27.44 & 27.53 & 26.91\\
AM-PRP & 97 & 86 & 69 & 53.52 & 59.89 & 56.06\\
AM-REC & 0 & 0 & 1 & 0.00 & 0.00 & 0.00\\
AM-TMP & 684 & 338 & 271 & 67.24 & 71.64 & 69.26\\
\bottomrule
\end{tabular}

\caption{10-fold cross-validation results for the PB-Br.v2 Adjudication set.}
\label{tab:adjudicacaoanotacaoplnbr}
\end{subtable}
}}
\end{table}

In the 10-fold cross-validation experiment on the Full set of the PB-Br.v2, we obtained F1 of 69.12 (Table \ref{tab:resultadofullplnbr}), which is between the 65.73 (Adjudication set) and 72, 71 (Agreement set) values.  We can speculate that despite the Adjudication set has more instances and results lower than the Agreement set, the latter supplies the system with syntactic trees without errors or trees that have frequent errors for which a standard treatment is predicted in the guidelines, and therefore can be identified by ML. It is also interesting to mention that the F1 difference between the system trained on the PB-Br.v1 gold-standard data and the system trained on the non-revised syntactic trees of the PB-Br.v2 is 10.48 F1 scores. The difference of around 10.0 F1 scores between a system trained on revised trees and on non-revised trees has already been investigated by \cite{palmeretal2005} and \cite{toutanovaetal2008} for the English language.

\begin{table}[!ht]
\center
\scriptsize
\scalebox{.88}{
\parbox{.50\linewidth}{
\begin{subtable}

\begin{tabular}{lrrrrrr}
\toprule
& \textbf{corr.} & \textbf{excess} & \textbf{missed} & \textbf{prec.} & \textbf{rec.} & \textbf{F1}\\
\textbf{Overall} & 11,047 & 5,335 & 4,537 & 67.45 & 70.89 & 69.12\\
\midrule
A0 & 2,703 & 671 & 500 & 80.10 & 84.37 & 82.15\\
A1 & 4,630 & 1,713 & 1,664 & 73.01 & 73.57 & 73.28\\
A2 & 898 & 974 & 702 & 48.07 & 56.09 & 51.67\\
A3 & 25 & 64 & 124 & 28.62 & 16.72 & 20.85\\
A4 & 18 & 10 & 30 & 68.00 & 36.07 & 45.21\\
AM-ADV & 159 & 150 & 139 & 50.00 & 52.37 & 50.64\\
AM-CAU & 54 & 69 & 86 & 43.71 & 39.87 & 40.99\\
AM-DIR & 0 & 4 & 13 & 0.00 & 0.00 & 0.00\\
AM-DIS & 183 & 116 & 161 & 60.95 & 52.57 & 56.20\\
AM-EXT & 72 & 41 & 54 & 63.05 & 57.06 & 59.51\\
AM-LOC & 529 & 460 & 198 & 53.73 & 72.96 & 61.75\\
AM-MNR & 302 & 352 & 254 & 46.42 & 54.34 & 49.84\\
AM-NEG & 360 & 42 & 58 & 88.91 & 85.90 & 87.23\\
AM-PRD & 37 & 123 & 134 & 24.33 & 20.79 & 21.96\\
AM-PRP & 147 & 99 & 75 & 59.75 & 66.54 & 62.48\\
AM-REC & 0 & 0 & 1 & 0.00 & 0.00 & 0.00\\
AM-TMP & 930 & 447 & 342 & 67.53 & 73.14 & 70.12\\
\bottomrule
\end{tabular}
\caption{10-fold cross-validation results for the Full set of the PB-Br.v2, comprising both Agreement and Adjudication sets, when using Alva-Manchego’s system.}
\label{tab:resultadofullplnbr}
\end{subtable}
}}
\quad\quad
\scalebox{.88}{
\parbox{.52\linewidth}{
\begin{subtable}

\begin{tabular}{lrrrrrr}
\toprule
& \textbf{corr.} & \textbf{excess} & \textbf{missed} & \textbf{prec.} & \textbf{rec.} & \textbf{F1}\\
\textbf{Overall} & 8,038 & 6,475 & 7,464 & 55.44 & 51.87 & 53.58\\
\midrule
A0 & 2,318 & 1,058 & 911 & 68.72 & 71.82 & 70.18\\
A1 & 3,497 & 2,979 & 2,755 & 54.13 & 55.93 & 55.00\\
A2 & 440 & 938 & 11,16 & 31.97 & 28.26 & 29.94\\
A3 & 0 & 6 & 150 & 0.00 & 0.00 & 0.00\\
A4 & 0 & 0 & 48 & 0.00 & 0.00 & 0.00\\
AM-ADV & 90 & 63 & 208 & 62.09 & 30.84 & 39.95\\
AM-CAU & 5 & 23 & 134 & 10.19 & 4.36 & 6.04\\
AM-DIR & 0 & 0 & 13 & 0.00 & 0.00 & 0.00\\
AM-DIS & 121 & 94 & 222 & 57.06 & 35.73 & 42.93\\
AM-EXT & 53 & 34 & 72 & 61.13 & 43.78 & 49.81\\
AM-LOC & 282 & 469 & 444 & 37.57 & 38.93 & 37.98\\
AM-MNR & 143 & 180 & 412 & 45.09 & 26.13 & 32.86\\
AM-NEG & 355 & 40 & 59 & 89.81 & 86.34 & 87.89\\
AM-PRD & 6 & 46 & 163 & 16.00 & 3.33 & 5.02\\
AM-PRP & 64 & 100 & 154 & 38.15 & 29.89 & 33.18\\
AM-REC & 0 & 0 & 1 & 0,00 & 0.00 & 0.00\\
AM-TMP & 664 & 445 & 602 & 59,92 & 52.51 & 55.83\\
\bottomrule
\end{tabular}
\caption{10-fold cross-validation results for the Full set of the PB-Br.v2 selection, comprising the Agreement and Adjudication sets, using Fonseca’s system.}
\label{tab:10foldplnbrfonseca}
\end{subtable}
}}
\end{table}

The contrast between the F1 values for the PB-Br.v1 and the PB-Br.v2 is noteworthy. Although the selection made on the PB-Br.v2  respects a distribution that should favor the SRL and this corpus is approximately 24\% larger than the PB-Br.v1, the results show a system with 10.48 F1 scores lower than the result obtained from the system trained by Alva-Manchego on the PB-Br.v1 corpus (F1 69.12 against 79.6). The evidence that the use of automatic parser without human revision adversely affects the performance of SRL systems become more evident when we contrast the results in Table \ref{tab:plnbrfullnopropbankbr} with those of Table \ref{tab:propbankbrnoplnbrfull}. The question, however, is whether this drop on performance actually represents a system with worse overall performance, or if the annotation in the automatic parser scenario is difficult to the point that the F1 value of the trained system in PB-Br.v2 (Table \ref{tab:resultadofullplnbr}) is considered good. To this end, we conducted two experiments: one annotating the PB-Br.v1 corpus with the system trained on the Full set of the PB-Br.v2 and another in the opposite direction – annotating the Full set of the PB-Br.v2 with the system trained on the PB-Br.v1 corpus.

We can see in Table \ref{tab:plnbrfullnopropbankbr} that the SRL in revised syntactic trees, using Alva-Manchego’s system trained on non-revised syntactic trees, is feasible, even though its quality is not comparable to that of a system trained on revised trees (F1 72.62 versus F1 79.6). We also noted that the value obtained of 72.62 is greater than the 10-fold cross-validation value of 69.12 of the system trained on the PB-Br.v2. Thus, we realized that in addition to a system trained on non-revised syntactic trees being able to annotate syntactic trees containing errors, it could annotate perfect trees with a performance superior than what it annotates in its own scenario. This means that training a SRL system on non-correct syntactic trees makes the system capture syntactic patterns of correct trees and learn to deal with the noise contained in not syntactically correct trees.

Doing the opposite direction of annotation, we can see in Table \ref{tab:propbankbrnoplnbrfull} that the quality of the SRL of the PB-Br.v2 problematic trees using the system trained on PB-Br.v1 is 14.36 F1 scores lower than the result of the 10-fold cross-validation for the Full set of PB-Br.v2 (F1 of 69.12 against 54.76). This result shows that a system trained on perfect syntactic trees faces greater difficulty when annotating faulty syntactic trees. As the PB-Br.v1 gold standard corpus does not contain noisy examples, a minimum deviation from the correct pattern of syntactic tree is enough to cause the system to make annotation errors.

\begin{table}[!ht]
\center
\scriptsize
\scalebox{.88}{
\parbox{.50\linewidth}{
\begin{subtable}{}
\begin{tabular}{lrrrrrr}
\toprule
& \textbf{corr.} & \textbf{excess} & \textbf{missed} & \textbf{prec.} & \textbf{rec.} & \textbf{F1}\\
\textbf{Overall} & 9,622 & 2,958 & 4,298 & 76.49 & 69.12 & 72.62\\
\midrule
A0 & 2,281 & 463 & 778 & 83.13 & 74.57 & 78.61\\
A1 & 4,187 & 863 & 1142 & 82.91 & 78.57 & 80.68\\
A2 & 772 & 526 & 613 & 59.48 & 55.74 & 57.55\\
A3 & 20 & 60 & 127 & 25.00 & 13.61 & 17.62\\
A4 & 35 & 26 & 79 & 57.38 & 30.70 & 40.00\\
A5 & 0 & 0 & 1 & 0.00 & 0.00 & 0.00\\
AM-ADV & 162 & 95 & 212 & 63.04 & 43.32 & 51.35\\
AM-CAU & 75 & 55 & 77 & 57.69 & 49.34 & 53.19\\
AM-DIR & 0 & 0 & 13 & 0.00 & 0.00 & 0.00\\
AM-DIS & 180 & 87 & 129 & 67.42 & 58.25 & 62.50\\
AM-EXT & 32 & 20 & 46 & 61.54 & 41.03 & 49.23\\
AM-LOC & 523 & 257 & 190 & 67.05 & 73.35 & 70.06\\
AM-MNR & 168 & 194 & 239 & 46.41 & 41.28 & 43.69\\
AM-NEG & 315 & 47 & 28 & 87.02 & 91.84 & 89.36\\
AM-PRD & 14 & 62 & 169 & 18.42 & 7.65 & 10.81\\
AM-PRP & 97 & 52 & 56 & 65.10 & 63.40 & 64.24\\
AM-REC & 0 & 0 & 8 & 0.00 & 0.00 & 0.00\\
AM-TMP & 761 & 151 & 388 & 83.44 & 66.23 & 73.85\\
\bottomrule
\end{tabular}

\caption{Annotation results for PB-Br.v1 using the SRL system trained on the Full set of the PB-Br.v2.}
\label{tab:plnbrfullnopropbankbr}
\end{subtable}
}}
\quad\quad
\scalebox{.88}{
\parbox{.52\linewidth}{
\begin{subtable}{}
\begin{tabular}{lrrrrrr}
\toprule
& \textbf{corr.} & \textbf{excess} & \textbf{missed} & \textbf{prec.} & \textbf{rec.} & \textbf{F1}\\
\textbf{Overall} & 7,972 & 5,559 & 7,612 & 58.92 & 51.16 & 54.76\\
\midrule
A0 & 2,113 & 1,040 & 1,090 & 67.02 & 65.97 & 66.49\\
A1 & 3,426 & 1,543 & 2,868 & 68.95 & 54.43 & 60.84\\
A2 & 478 & 526 & 1,122 & 47.61 & 29.88 & 36.71\\
A3 & 11 & 24 & 138 & 31.43 & 7.38 & 11.96\\
A4 & 8 & 13 & 40 & 38.10 & 16.67 & 23.19\\
A5 & 0 & 0 & 0 & -- & -- & --\\
AM-ADV & 100 & 179 & 198 & 35.84 & 33.56 & 34.66\\
AM-CAU & 36 & 106 & 104 & 25.35 & 25.71 & 25.53\\
AM-DIR & 0 & 2 & 13 & 0.00 & 0.00 & 0.00\\
AM-DIS & 106 & 74 & 238 & 58.89 & 30.81 & 40.46\\
AM-EXT & 63 & 45 & 63 & 58.33 & 50.00 & 53.85\\
AM-LOC & 412 & 520 & 315 & 44.21 & 56.67 & 49.67\\
AM-MNR & 242 & 472 & 314 & 33.89 & 43.53 & 38.11\\
AM-NEG & 204 & 49 & 214 & 80.63 & 48.80 & 60.80\\
AM-PRD & 8 & 245 & 163 & 3.16 & 4.68 & 3.77\\
AM-PRP & 101 & 126 & 121 & 44.49 & 45.50 & 44.99\\
AM-REC & 0 & 25 & 1 & 0.00 & 0.00 & 0.00\\
AM-TMP & 664 & 570 & 608 & 53.81 & 52.20 & 52.99\\
\bottomrule
\end{tabular}

\caption{Annotation results for the Full set of PB-Br.v2 using the SRL system trained on PB-Br.v1.}
\label{tab:propbankbrnoplnbrfull}
\end{subtable}
}}
\end{table}

To strengthen the analysis of results, we annotated the PB-Br.v2 Agreement and Adjudication sets with the system trained on the PB-Br.v1. Table \ref{tab:propbankbrnoconcordancia} shows the result of 58.96 F1 in the annotation of the Agreement set of the PB-Br.v2 by the system trained in PB-Br.v1 and Table \ref{tab:propbankbrnoadjudicado} presents F1 of 52.66 for annotation of the Adjudication set. The difference in the performance of annotation of the sets was 6.3 F1 scores, while the difference of the 10-fold cross-validation performed on these sets was 6.98 F1 scores. As the difference in the performance of annotation of the sets is close, we understand that, indeed, there is a syntactic complexity threshold that distinguishes the  Agreement and the Adjudication sets of the PB-Br.v2. 
Table \ref{tab:confusionmatrix2} and Table \ref{tab:confusionmatrix1} show confusion matrices for the results presented on Table \ref{tab:plnbrfullnopropbankbr} and Table \ref{tab:propbankbrnoplnbrfull}. 
These matrices show that our SRL system trained on noisy data performs better when annotating revised trees (Table 9) than the system trained on revised trees when annotating noisy data (Table 10). For example, there are 32\% of AM-TMP syntactic nodes not identified by the SRL trained on PB-Br.v.1 and tested on noisy data (PB-Br.v2) whereas there are only 6\% of AM-TMP not identified by the SRL trained on PB-Br.v2 and tested on the treebank (PB.Br-v1).

Finally, we conducted a 10-fold cross-validation experiment with Fonseca’s system and the PB-Br.v2 full. The F1 value of 53,58 for the system that does not use syntactic features presented in Table \ref{tab:10foldplnbrfonseca} shows that even with the use of syntactic trees with errors, the SRL approach for syntactic trees is better.

\begin{table}[!ht]
\center
\scriptsize
\scalebox{.88}{
\parbox{.50\linewidth}{
\begin{subtable}{}
\begin{tabular}{lrrrrrr}
\toprule
& \textbf{corr.} & \textbf{excess} & \textbf{missed} & \textbf{prec.} & \textbf{rec.} & \textbf{F1}\\
\textbf{Overall} & 2,803 & 1,718 & 2,184 & 62.00 & 56.21 & 58.96\\
\midrule
A0 & 631 & 372 & 322 & 62.91 & 66.21 & 64.52\\
A1 & 1,455 & 491 & 998 & 74.77 & 59.32 & 66.15\\
A2 & 210 & 144 & 380 & 59.32 & 35.59 & 44.49\\
A3 & 4 & 10 & 49 & 28.57 & 7.55 & 11.94\\
A4 & 4 & 4 & 13 & 50.00 & 23.53 & 32.00\\
AM-ADV & 11 & 58 & 11 & 15.94 & 50.00 & 24.18\\
AM-CAU & 5 & 24 & 13 & 17.24 & 27.78 & 21.28\\
AM-DIR & 0 & 1 & 0 & 0.00 & 0.00 & 0.00\\
AM-DIS & 11 & 27 & 11 & 28.95 & 50.00 & 36.67\\
AM-EXT & 14 & 14 & 13 & 50.00 & 51.85 & 50.91\\
AM-LOC & 92 & 144 & 61 & 38.98 & 60.13 & 47.30\\
AM-MNR & 73 & 130 & 64 & 35.96 & 53.28 & 42.94\\
AM-NEG & 81 & 14 & 72 & 85.26 & 52.94 & 65.32\\
AM-PRD & 0 & 58 & 1 & 0.00 & 0.00 & 0.00\\
AM-PRP & 27 & 44 & 29 & 38.03 & 48.21 & 42.52\\
AM-REC & 0 & 6 & 0 & 0.00 & 0.00 & 0.00\\
AM-TMP & 185 & 177 & 132 & 51.10 & 58.36 & 54.49\\
\bottomrule
\end{tabular}
\caption{Annotation results for the PB-Br.v2 Agreement set.}
\label{tab:propbankbrnoconcordancia}
\end{subtable}
}}
\quad\quad
\scalebox{.88}{
\parbox{.52\linewidth}{
\begin{subtable}{}
\begin{tabular}{lrrrrrr}
\toprule
& \textbf{corr.} & \textbf{excess} & \textbf{missed} & \textbf{prec.} & \textbf{rec.} & \textbf{F1}\\
\textbf{Overall} & 5,166 & 3,844 & 5,444 & 57.34 & 48.69 & 52.66\\
\midrule
A0 & 1,482 & 668 & 768 & 68.93 & 65.87 & 67.36\\
A1 & 1,972 & 1,051 & 1,873 & 65.23 & 51.29 & 57.43\\
A2 & 264 & 386 & 742 & 40.62 & 26.24 & 31.88\\
A3 & 7 & 14 & 89 & 33.33 & 7.29 & 11.97\\
A4 & 4 & 9 & 27 & 30.77 & 12.90 & 18.18\\
AM-ADV & 89 & 121 & 187 & 42.38 & 32.25 & 36.63\\
AM-CAU & 31 & 82 & 91 & 27.43 & 25.41 & 26.38\\
AM-DIR & 0 & 1 & 13 & 0.00 & 0.00 & 0.00\\
AM-DIS & 95 & 47 & 227 & 66.90 & 29.50 & 40.95\\
AM-EXT & 49 & 31 & 50 & 61.25 & 49.49 & 54.75\\
AM-LOC & 320 & 376 & 254 & 45.98 & 55.75 & 50.39\\
AM-MNR & 169 & 342 & 250 & 33.07 & 40.33 & 36.34\\
AM-NEG & 123 & 35 & 142 & 77.85 & 46.42 & 58.16\\
AM-PRD & 8 & 187 & 162 & 4.10 & 4.71 & 4.38\\
AM-PRP & 74 & 82 & 92 & 47.44 & 44.58 & 45.96\\
AM-REC & 0 & 19 & 1 & 0.00 & 0.00 & 0.00\\
AM-TMP & 479 & 393 & 476 & 54.93 & 50.16 & 52.44\\
\bottomrule
\end{tabular}
\caption{Annotation results for the PB-Br.v2 Adjudication set.}
\label{tab:propbankbrnoadjudicado}
\end{subtable}
}}
\end{table}

\begin{table}[!ht]
\center
\scriptsize
\scalebox{.84}{
\begin{tabular}{r|lrrrrrrrrrrrrrrrrrrr}
\toprule
& & -1 & 0 & 1 & 2 & 3 & 4 & 5 & 6 & 7 & 8 & 9 & 10 & 11 & 12 & 13 & 14 & 15 & 16 & 17\\
\midrule
-1 & -NONE- & 0 & 615 & 605 & 168 & 26 & 21 & 0 & 110 & 31 & 1 & 81 & 4 & 81 & 70 & 24 & 81 & 28 & 2 & 186\\
0 & A0 & 130 & 2,281 & 265 & 33 & 2 & 1 & 1 & 0 & 9 & 0 & 1 & 4 & 1 & 5 & 0 & 6 & 0 & 1 & 3\\
1 & A1 & 292 & 140 & 4,187 & 247 & 26 & 21 & 0 & 8 & 5 & 2 & 5 & 11 & 20 & 33 & 0 & 9 & 5 & 5 & 33\\
2 & A2 & 76 & 10 & 196 & 772 & 48 & 25 & 0 & 5 & 9 & 5 & 4 & 5 & 36 & 50 & 0 & 17 & 14 & 0 & 26\\
3 & A3 & 5 & 1 & 19 & 22 & 20 & 1 & 0 & 0 & 0 & 1 & 0 & 2 & 1 & 4 & 0 & 1 & 0 & 0 & 3\\
4 & A4 & 6 & 0 & 2 & 3 & 2 & 35 & 0 & 0 & 1 & 2 & 0 & 0 & 0 & 4 & 0 & 0 & 3 & 0 & 3\\
5 & A5 & 0 & 0 & 0 & 0 & 0 & 0 & 0 & 0 & 0 & 0 & 0 & 0 & 0 & 0 & 0 & 0 & 0 & 0 & 0\\
6 & AM-ADV & 31 & 4 & 1 & 1 & 0 & 0 & 0 & 162 & 1 & 0 & 18 & 3 & 1 & 14 & 3 & 4 & 1 & 0 & 13\\
7 & AM-CAU & 16 & 0 & 1 & 8 & 3 & 1 & 0 & 7 & 75 & 0 & 1 & 0 & 3 & 10 & 0 & 2 & 1 & 0 & 2\\
8 & AM-DIR & 0 & 0 & 0 & 0 & 0 & 0 & 0 & 0 & 0 & 0 & 0 & 0 & 0 & 0 & 0 & 0 & 0 & 0 & 0\\
9 & AM-DIS & 34 & 0 & 1 & 3 & 0 & 0 & 0 & 19 & 3 & 0 & 180 & 0 & 3 & 2 & 0 & 4 & 2 & 0 & 16\\
10 & AM-EXT & 7 & 0 & 1 & 2 & 0 & 0 & 0 & 4 & 0 & 0 & 2 & 32 & 0 & 2 & 0 & 0 & 0 & 0 & 2\\
11 & AM-LOC & 43 & 0 & 24 & 69 & 3 & 6 & 0 & 7 & 3 & 1 & 3 & 0 & 523 & 18 & 0 & 26 & 0 & 0 & 54\\
12 & AM-MNR & 48 & 2 & 9 & 22 & 7 & 2 & 0 & 39 & 5 & 1 & 5 & 7 & 12 & 168 & 1 & 11 & 1 & 0 & 22\\
13 & AM-NEG & 30 & 0 & 0 & 0 & 0 & 0 & 0 & 0 & 1 & 0 & 1 & 0 & 0 & 0 & 315 & 0 & 0 & 0 & 15\\
14 & AM-PRD & 20 & 1 & 1 & 4 & 0 & 0 & 0 & 2 & 4 & 0 & 3 & 1 & 7 & 9 & 0 & 14 & 0 & 0 & 10\\
15 & AM-PRP & 6 & 0 & 7 & 22 & 7 & 0 & 0 & 2 & 3 & 0 & 2 & 0 & 1 & 0 & 0 & 2 & 97 & 0 & 0\\
16 & AM-REC & 0 & 0 & 0 & 0 & 0 & 0 & 0 & 0 & 0 & 0 & 0 & 0 & 0 & 0 & 0 & 0 & 0 & 0 & 0\\
17 & AM-TMP & 55 & 1 & 9 & 9 & 3 & 1 & 0 & 9 & 2 & 0 & 3 & 9 & 24 & 18 & 0 & 6 & 1 & 0 & 761\\
\bottomrule
\end{tabular}
}
\caption{Confusion Matrix of SRL on PB-Br.v1 using system trained on PB-Br.v2.}
\label{tab:confusionmatrix2}
\end{table}

\begin{table}[!ht]
\center
\scriptsize
\scalebox{.84}{
\begin{tabular}{r|lrrrrrrrrrrrrrrrrrrr}
\toprule
& & -1 & 0 & 1 & 2 & 3 & 4 & 5 & 6 & 7 & 8 & 9 & 10 & 11 & 12 & 13 & 14 & 15 & 16 & 17\\
\midrule
-1 & -NONE- & 0 & 961 & 2,242 & 564 & 61 & 16 & 143 & 76 & 4 & 206 & 2 & 20 & 231 & 193 & 195 & 101 & 90 & 1 & 485\\
 0 & A0 & 600 & 2,113 & 365 & 46 & 3 & 1 & 2 & 6 & 1 & 2 & 0 & 1 & 1 & 4 & 1 & 6 & 0 & 0 & 1\\
 1 & A1 & 1,139 & 93 & 3,426 & 236 & 10 & 2 & 1 & 0 & 1 & 0 & 0 & 19 & 8 & 11 & 0 & 5 & 1 & 0 & 17\\
 2 & A2 & 307 & 11 & 96 & 478 & 28 & 7 & 0 & 1 & 0 & 0 & 0 & 0 & 27 & 16 & 0 & 9 & 11 & 0 & 13\\
 3 & A3 & 11 & 2 & 3 & 6 & 11 & 1 & 0 & 0 & 0 & 1 & 0 & 0 & 0 & 0 & 0 & 0 & 0 & 0 & 0\\
 4 & A4 & 5 & 0 & 3 & 2 & 0 & 8 & 0 & 0 & 0 & 0 & 0 & 0 & 0 & 1 & 0 & 0 & 0 & 0 & 2\\
 5 & AM-ADV & 114 & 1 & 6 & 1 & 0 & 0 & 100 & 2 & 0 & 8 & 0 & 3 & 3 & 14 & 2 & 11 & 3 & 0 & 11\\
 6 & AM-CAU & 76 & 4 & 7 & 6 & 1 & 1 & 1 & 36 & 1 & 0 & 0 & 0 & 1 & 4 & 0 & 3 & 0 & 0 & 1\\
 7 & AM-DIR & 1 & 0 & 0 & 1 & 0 & 0 & 0 & 0 & 0 & 0 & 0 & 0 & 0 & 0 & 0 & 0 & 0 & 0 & 0\\
 8 & AM-DIS & 60 & 0 & 1 & 1 & 0 & 0 & 3 & 0 & 0 & 106 & 0 & 0 & 2 & 6 & 0 & 0 & 0 & 0 & 1\\
 9 & AM-EXP & 0 & 0 & 0 & 0 & 0 & 0 & 0 & 0 & 0 & 0 & 0 & 0 & 0 & 0 & 0 & 0 & 0 & 0 & 0\\
10 & AM-EXT & 33 & 0 & 6 & 2 & 1 & 0 & 1 & 0 & 0 & 0 & 0 & 63 & 0 & 1 & 0 & 1 & 0 & 0 & 0\\
11 & AM-LOC & 271 & 3 & 30 & 97 & 6 & 2 & 5 & 4 & 0 & 7 & 0 & 2 & 412 & 24 & 1 & 9 & 6 & 0 & 53\\
12 & AM-MNR & 237 & 1 & 32 & 72 & 14 & 2 & 24 & 8 & 4 & 3 & 0 & 13 & 24 & 242 & 1 & 11 & 5 & 0 & 21\\
13 & AM-NEG & 40 & 0 & 8 & 1 & 0 & 0 & 0 & 0 & 0 & 0 & 0 & 0 & 0 & 0 & 204 & 0 & 0 & 0 & 0\\
14 & AM-PRD & 169 & 10 & 17 & 24 & 0 & 3 & 2 & 5 & 0 & 1 & 0 & 0 & 5 & 7 & 0 & 8 & 0 & 0 & 2\\
15 & AM-PRP & 74 & 0 & 0 & 37 & 6 & 3 & 2 & 1 & 1 & 0 & 0 & 0 & 0 & 1 & 0 & 0 & 101 & 0 & 1 \\
16 & AM-REC & 4 & 1 & 17 & 2 & 0 & 0 & 0 & 0 & 0 & 1 & 0 & 0 & 0 & 0 & 0 & 0 & 0 & 0 & 0\\
17 & AM-TMP & 397 & 3 & 35 & 24 & 8 & 2 & 14 & 1 & 1 & 9 & 0 & 5 & 13 & 32 & 14 & 7 & 5 & 0 & 664\\
\bottomrule
\end{tabular}
}
\caption{Confusion Matrix of SRL on PB-Br.v2 using system trained on PB-Br.v1.}
\label{tab:confusionmatrix1}
\end{table}

\section{Conclusions}
\label{sec:conclusions}

The obtained results show that a SRL system responsive to syntactic errors, trained on noisy data (non-revised syntactic trees), performs a better SRL on noisy data than when it is trained on revised trees (treebank). 
We also noted that the SRL system obtained better results when tested on the set in which there was full agreement between annotators and lower results when tested on the set in which there was disagreement between annotators. 
We did not include in PB-Br.v2 any annotation mark to distinguish well-formed parse trees from those containing parsing errors. For this reason, it was not possible to verify whether the existence of parsing errors are correlated to the drop in inter-annotator agreement rates. During the adjudication process, however, we realized that this correlation probably exists and it would be worthwhile to explore this hypothesis in future work. 
Based on confusion matrices' results, however, we speculate that the learning from non-revised trees allows the system to better identify the SRL candidates. 
We also noted that Fonseca’s system \cite{fonseca2013} performs a SRL inferior to Alva-Manchego’s  \cite{alvamanchego2013}, even in an unfavorable scenario for the latter system. We believe that, for Brazilian Portuguese, the use of syntactic trees for the SRL task is still the most promising means and, therefore, efforts to the improvement of syntactic parsers are welcome. We also understand that in a real scenario of application (on-the-fly processing), the data is passed directly from the syntactic parser to the SRL system without human intervention to correct the trees. Therefore, we can say that in a real scenario of application, the training of a SRL system on non-revised syntactic trees corpus, such as the PB-Br.v2, provides better annotation of journalistic texts than the system trained on revised syntactic trees corpus (PB-Br.v1).
\section*{Acknowledgments}

Part of the research developed for this work was sponsored by \emph{Samsung Eletr\^onica da Amaz\^onia Ltda.} under the terms of Brazilian federal law number 8.248/91. Part of the results presented in this paper were obtained through research  activity in the project titled ``Semantic Processing of Brazilian Portuguese Texts'',  sponsored by \emph{Samsung Eletr\^onica da Amaz\^onia Ltda.} under the terms of Brazilian federal law number 8.248/91.

\bibliographystyle{splncs03}
\bibliography{referencias}

\end{document}